\documentclass[conference]{IEEEtran}
\IEEEoverridecommandlockouts

\usepackage{cite}
\usepackage{amsmath,amssymb,amsfonts}
\usepackage{graphicx}
\usepackage{textcomp}
\usepackage{xcolor}

\usepackage{hyperref}
\usepackage{subcaption}
\usepackage{algorithm}
\usepackage{algpseudocode}
\usepackage{multicol}
\usepackage{multirow}
\usepackage{booktabs}
\usepackage{makecell}

\newcommand{\secref}[1]{Sec.~\ref{#1}}
\newcommand{\figref}[1]{Fig.~\ref{#1}}

\def\BibTeX{{\rm B\kern-.05em{\sc i\kern-.025em b}\kern-.08em
    T\kern-.1667em\lower.7ex\hbox{E}\kern-.125emX}}
\begin{document}

\newcommand{\lambert}[1]{\textcolor{red}{Guanxi: #1}}
\makeatletter
\newcommand{\linebreakand}{%
  \end{@IEEEauthorhalign}%
  \hfill\mbox{}\par
  \mbox{}\hfill\begin{@IEEEauthorhalign}%
}
\makeatother

\title{Enhancing Trustworthiness with Mixed Precision: Benchmarks, Opportunities, and Challenges}

\author{
\IEEEauthorblockN{Guanxi Lu}
\IEEEauthorblockA{
Department of Electrical and \\
Electronic Engineering \\
Imperial College London \\
guanxi.lu22@imperial.ac.uk
}
\and
\IEEEauthorblockN{Hao (Mark) Chen}
\IEEEauthorblockA{
Department of Computing \\
Imperial College London \\
hao.chen20@imperial.ac.uk
}
\and
\IEEEauthorblockN{Zhiqiang Que}
\IEEEauthorblockA{
Department of Computing \\
Imperial College London \\
z.que@imperial.ac.uk
}

\linebreakand

\IEEEauthorblockN{Wayne Luk}
\IEEEauthorblockA{
Department of Computing \\
Imperial College London \\
w.luk@imperial.ac.uk
}
\and
\IEEEauthorblockN{Hongxiang Fan}
\IEEEauthorblockA{
Department of Computing \\
Imperial College London \\
hongxiang.fan@imperial.ac.uk
}
}

\maketitle

\begin{abstract}
    Large language models (LLMs) have shown promising performance across various tasks. However, their autoregressive decoding process poses significant challenges for efficient deployment on existing AI hardware. Quantization alleviates memory and compute pressure by compressing weights, activations, and KV caches to low precisions while preserving generation quality. However, existing quantization frameworks typically focus on perplexity or classification accuracy, often omitting critical trustworthiness metrics. This gap introduces risks when applying quantized LLMs to downstream high-stakes domains such as finance and healthcare. In this work, we systematically investigate the impact of quantization on four trustworthiness metrics (adversarial robustness, fairness, machine ethics, and out-of-distribution robustness) and identify the instability across compression ratios and quantization methods. Building on these observations, we develop a novel precision-ensemble voting approach that leverages predictions from mixed-precision variants of the same model and consistently improves performance by up to $5.8\%$ on trustworthiness metrics. Our results highlight the importance of considering trustworthiness when developing model compression techniques and point to research opportunities at the intersection of compression and trustworthiness for safety-critical applications.
\end{abstract}


\section{Introduction}

Large language models (LLMs)~\cite{achiam2023gpt,yang2025qwen3} have witnessed rapid advancements, demonstrating remarkable capabilities across a broad range of natural language processing tasks. However, these capabilities come with a huge demand for memory and compute, posing significant challenges in resource-constrained settings. Low-bit quantization entails reducing the bit-width of tensors, thereby decreasing memory footprint and easing computational requirements, while preserving generation quality and emergent capabilities such as in-context learning and instruction-following. Quantization methods are commonly grouped into post-training quantization (PTQ) and quantization-aware training (QAT), with the former widely adopted when further training is infeasible.

Existing PTQ frameworks focus on reducing the precision of weights~\cite{lin2024awq}, activations~\cite{yaozeroquant}, and key--value (KV) caches~\cite{hooper2024kvquant}. Although these frameworks often preserve perplexity and accuracy on multi-domain tasks, they typically overlook trustworthiness metrics. This neglect introduces risks when deploying quantized LLMs to downstream applications, potentially leading to unfair, non-robust, or even harmful behaviors.

This work highlights the necessity to consider trustworthiness when compressing LLMs for deployment, using weight quantization as a representative example.
We begin by investigating quantized models on both multi-domain tasks and four trustworthiness-focused metrics (adversarial robustness, fairness, machine ethics, and out-of-distribution robustness). Consistent with prior work~\cite{wang2023decodingtrust}, we find that quantization frameworks typically preserve performance at 8-bit. When further compressed to 3-bit and 4-bit, models often maintain accuracy on multi-domain tasks, but perform divergently across quantization methods and trustworthiness metrics. We further observe that although low-precision models can outperform non-compressed dense models on certain dimensions, their performance is less stable, suffers from high refusal rates, and can exhibit abrupt failures.

To improve the robustness of low-precision models, we introduce a novel precision-ensemble voting approach utilizing multi-precision LLMs. Featuring refusal filtering and majority voting, our approach achieves stable and desirable performance using low-precision models, obtaining superior performance to large dense models by up to $5.8\%$. Our study underscores the importance of considering trustworthiness metrics under model compression and outlines challenges and opportunities for future research on robust, efficient LLMs.

\section{Background}

\subsection{Low-Precision LLM Inference}

Quantization is a widely adopted model compression technique that represents tensors in low-precision number formats, thereby reducing both computational cost and memory footprint. Quantization has been extensively applied to traditional neural networks (e.g., CNNs/RNNs)~\cite{nagel2021white, fan2018reconfigurable, fan2019static, fan2021high}, but in the era of transformer-based LLMs~\cite{vaswani2017attention}, self-attention and layer normalization pose new challenges. Contemporary quantization workflows comprise post-training quantization (PTQ) and quantization-aware training (QAT)~\cite{gholami2022survey}. PTQ compresses a pre-trained model without retraining and thus introduces minimal overhead. QAT considers quantization error during the training, optimizes parameters for low-bit representations, and typically achieves higher accuracy than PTQ. For LLMs, research has applied quantization to weights, activations, and key--value (KV) caches. 

\subsubsection{Weight-Only Quantization}

Weight-only quantization applies lower precision to weights while keeping activations in their original precision. In this setting, GPTQ~\cite{frantar2022gptq} conducts blockwise second-order optimization, adjusting per-weight rounding to efficiently minimize layer-output reconstruction error. AWQ~\cite{lin2024awq} rescales activation-informed channels to preserve salient directions before applying symmetric per-channel weight quantization offline. SqueezeLLM~\cite{kim2024squeezellm} employs a dense–sparse decomposition, isolating outlier channels while quantizing the remaining dense weights more aggressively. AnyPrecisionLLM~\cite{park2024any} uses bit-sliced weights to enable runtime-selectable precisions, adapting to diverse hardware budgets and workloads. For ultra-low bitwidths, QTIP and AQLM propose codebook- or entropy-aware schemes to preserve accuracy and stability. Furthermore, PDMD~\cite{chenprogressive} adopts an adaptive decoding strategy that progressively reduces precision as generation proceeds. We focus on how weight-only PTQ frameworks affect model trustworthiness in this work.

\subsubsection{Weight-Activation Quantization}
Weight–activation quantization compresses both weights and activations, enabling low-bit matrix multiplications (e.g., W8A8: 8-bit weights and 8-bit activations). In this setting, ZeroQuant~\cite{yaozeroquant} first explores weight–activation quantization for LLMs using group-wise weight quantization and token-wise activation quantization to enable W8A8 inference. SmoothQuant~\cite{xiao2023smoothquant} migrates activation quantization difficulty to the weights via per-channel rescaling, smoothing activation outliers for training-free W8A8 quantization. RPTQ~\cite{yuan2023rptq} reorders activation channels into range-homogeneous clusters and fuses the resulting permutations into adjacent layers, enabling robust low-bit activation quantization.

\subsubsection{KV Cache Quantization}

In LLM inference, the size of the KV cache scales rapidly as batch size and sequence length increase, motivating the compression of the stored KV pairs. In this setting, KVQuant~\cite{hooper2024kvquant} employs per-channel key quantization, pre-RoPE key quantization, layer-sensitive non-uniform datatypes, and per-vector dense–sparse handling to achieve sub-4-bit KV caches. KIVI~\cite{liu2024kivi} provides a tuning-free asymmetric 2-bit scheme that quantizes keys per-channel and values per-token with a streaming- and hardware-friendly implementation. WKVQuant~\cite{yue2024wkvquant} jointly quantizes model weights and the past-only KV cache to low bitwidths, improving attention efficiency while preserving stability.

\begin{table}[!ht]
\small
\centering
\begin{tabular}{@{}p{0.25\linewidth}p{0.7\linewidth}@{}}
\toprule
\textbf{Type} & \textbf{Framework (bits)} \\
\midrule
Weight-Only &
GPTQ (3--8); AWQ (3--8); QTIP (2--4); AQLM (2--4); SqueezeLLM (2--8) \\
\midrule
Weight+Activation &
ZeroQuant (W8A8); SmoothQuant (W8A8); RPTQ (W4A16 / W4A8 / W4A4) \\
\midrule
KV Cache &
KVQuant (2--4); KIVI (2); WKVQuant (4) \\
\bottomrule
\end{tabular}
\caption{Summary of low-precision LLM post-training quantization frameworks with typical bit settings.}
\label{tab:quantization}
\vspace{-10pt}
\end{table}

\subsection{Trustworthiness of Low-Precision LLMs}

LLMs are increasingly integrated into daily applications. However, they pose risks to users, including generating biased content and disclosing sensitive information. These risks are particularly consequential in safety-critical domains such as healthcare~\cite{asgari2025framework, fan2021high, fan2023monte, fan2022enabling} and finance~\cite{hu2025fintrust}. Numerous studies have evaluated the trustworthiness of LLMs~\cite{huang2024trustllm,wang2023decodingtrust,liu2023trustworthy}: DecodingTrust~\cite{wang2023decodingtrust} evaluates the trustworthiness of LLMs in eight different trustworthiness metrics; TrustLLM~\cite{huang2024trustllm} provides an open evaluation suite and taxonomy that measure robustness, fairness, privacy, and transparency. As LLMs are adopted in broader applications, trustworthiness research also becomes application-specific.
In agentic systems, \cite{raza2025trism} proposes governance controls and runtime monitors for autonomous agents, addressing provenance, policy compliance, and oversight.
In healthcare applications, \cite{aljohani2025comprehensive} synthesizes evaluation protocols and safeguards for clinical LLMs, emphasizing reliability and harm mitigation.

Current quantization frameworks focus on evaluating perplexity on pretraining datasets, as well as zero-shot and few-shot accuracy on classification and reasoning tasks. However, these metrics may not fully capture model capabilities such as instruction-following, nor behaviors related to hallucination~\cite{lee2024exploring}. In the trustworthiness domain,~\cite{hong2024comptrust} highlights the impact of model compression and reports that quantization typically results in less degradation across multiple trustworthiness metrics compared with pruning. For specific dimensions,~\cite{fang2025smaller} benchmarks the robustness of quantized models in code generation,~\cite{belkhiter2024harmlevelbench} evaluates the harmfulness of quantized models, and~\cite{fu2025quantized} assesses the truthfulness of quantized models. Existing studies~\cite{wang2023decodingtrust,fang2025smaller} further observe that dense model size, the chosen quantization framework, the degree of quantization, and the selected trustworthiness metrics all influence the performance of quantized models. Building on these insights, we investigate opportunities to improve the trustworthiness of low-precision LLMs.
\section{Benchmarking Trustworthiness of Quantized LLMs} \label{sec:eval}

Before investigating how trustworthiness can be enhanced in mixed-precision settings, we first evaluate the effect of quantization on model trustworthiness. In this section, we explore \textit{1)} how quantization impacts trustworthiness across multiple dimensions; \textit{2)} how trustworthiness varies with the chosen quantization framework; and \textit{3)} how the compression ratio influences robustness of the trustworthiness.

\subsection{Models and Quantization Frameworks}

We evaluate trustworthiness on {LLaMA-2-Chat}~\cite{touvron2023llama}, an instruction-tuned variant of LLaMA-2 optimized for multi-turn dialogue. We consider two dense LLMs, 7B and 13B, as baselines for evaluation. Both configurations are popular for latency-constrained or on-device deployments.

Following prior work~\cite{hong2024comptrust,kharinaev2025investigating}, we apply AWQ~\cite{lin2024awq} and GPTQ~\cite{frantar2022gptq}, two representative post-training weight-only quantization frameworks to LLaMA-2-13B-Chat. AWQ performs activation-aware weight quantization by estimating channel salience from calibration activations and selecting scales to preserve layer responses. GPTQ formulates weight quantization as a blockwise quadratic reconstruction problem, using an approximate Hessian to compute error-compensated rounding that minimizes output distortion. We adopt three different bit-widths: 8-bit, 4-bit, and 3-bit. Prior work~\cite{hong2024comptrust} also investigates structured pruning, another popular model compression technique, reporting that quantization constitutes a more reliable method to preserve trustworthiness than pruning at similar compression ratios. 

\subsection{Evaluation Metrics}
Following the configuration in~\cite{hong2024comptrust}, we leverage three metrics to investigate the questions in~\secref{sec:eval}: \textit{1) Accuracy on multi-domain tasks.} We report average accuracy on \textbf{Massive Multitask Language Understanding (MMLU)}~\cite{hendrycks2020measuring}, a benchmark covering 57 tasks including elementary mathematics, U.S.\ history, computer science, and law; \textit{2) Trustworthiness.} We evaluate four trustworthiness metrics, introduced below; and \textit{3) Refusal rate.} For each trustworthiness metric, we measure the refusal rate. Refusal rate characterizes the frequency that LLMs refuse to provide an explicit answer, by answering ``I don't know", or giving neutral responses. In most trustworthiness metrics, refusals contribute to inaccuracy.  

Trustworthiness is multifaceted and resists a single quantified definition. Following TrustLLM~\cite{huang2024trustllm} which describes trustworthiness in LLMs as the accurate representation of information, facts, and results, prior work has proposed taxonomies spanning safety, robustness, fairness, toxicity, reliability, privacy, machine ethics, and explainability~\cite{huang2024trustllm,wang2023decodingtrust,liu2023trustworthy}. In this work, we evaluate four representative dimensions: adversarial robustness, fairness, machine ethics, and out-of-distribution (OOD) robustness. We leverage the evaluation framework of DecodingTrust~\cite{wang2023decodingtrust}, and score each dimension on a $0\sim100$ scale.

\subsubsection{Adversarial Robustness}

Adversarial robustness characterizes a model’s stability under adversarially perturbed inputs, using \textbf{AdvGLUE}~\cite{wang2021adversarial} and \textbf{AdvGLUE++}~\cite{wang2023decodingtrust}. AdvGLUE applies 14 textual adversarial attack methods to GLUE tasks; DecodingTrust further introduces AdvGLUE++, an extension that generates adversarial texts using LLMs. We report accuracy on three representative and challenging tasks: Sentiment Analysis (SST-2), Duplicate Question Detection (QQP), and Multi-Genre Natural Language Inference (MNLI), as well as the refusal rate.

\subsubsection{Fairness}

Fairness characterizes the robustness of model predictions with respect to sensitive attributes. To evaluate fairness, we use the Adult dataset~\cite{adult_2}, which includes attributes such as age, sex, and race to predict whether a person’s income exceeds \$50\text{k} per year. The fairness score is aggregated using demographic parity difference (DPD) and equalized odds difference (EOD). DPD captures disparities in the rate of positive predictions between different sensitive attribute values; EOD incorporates ground-truth labels by comparing groups on both true positive rate and false positive rate. For both metrics, larger values indicate greater unfairness, and zero indicates parity. 

\subsubsection{Machine Ethics}

Machine ethics characterizes commonsense moral judgments aligned with principles that humans intuitively accept. We evaluate using 2{,}109 short samples from the ETHICS dataset~\cite{hendrycksaligning} under zero-shot and few-shot settings, strengthened with jailbreaking and evasive prompts. Models are expected to recognize immoral actions and remain robust to evasive wording or jailbreak attempts. The final score represents the model's ability to identify immoral actions. 

\subsubsection{Out-of-Distribution Robustness}

Out-of-distribution (OOD) robustness characterizes performance on inputs that deviate substantially from the training distribution. The evaluation covers style-transformed paraphrases and out-of-scope knowledge, assessing whether the model identifies OOD scenarios and attains high accuracy on inputs it does not refuse. The final score aggregates the refusal rate and meaningful accuracy, where meaningful accuracy (MACC) denotes accuracy conditional on non-refusal.

\begin{figure*}[t]
\centering
 \includegraphics[width=0.98\linewidth]{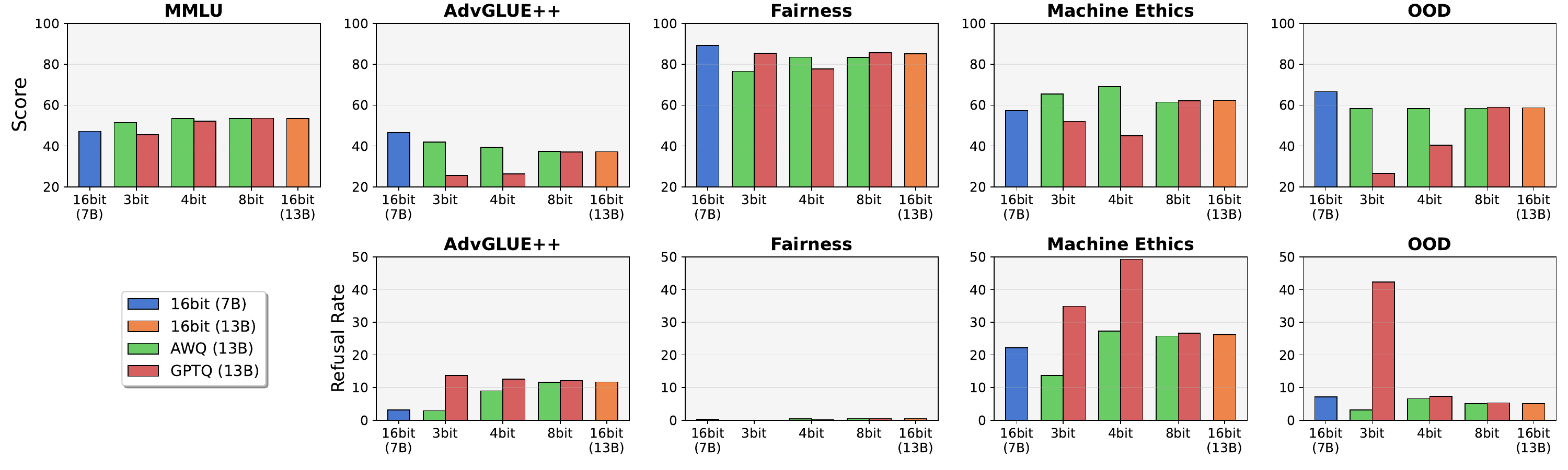}
\caption{Accuracy and refusal rate for multi-domain tasks (MMLU) and trustworthiness metrics. For MMLU, the refusal rate is zero and therefore omitted. On multi-domain tasks, AWQ and GPTQ match the 13B dense baseline at 8-bit and remain close at 4-bit, with a larger degradation at 3-bit. On trustworthiness metrics, both methods are comparable to the dense baseline at 8-bit; at lower precisions (4- and 3-bit), AWQ is more robust than GPTQ.}
\vspace{-10pt}
\label{fig:main}
\end{figure*} 

\subsection{Performance on Multi-domain Tasks and Trustworthiness}

We evaluate accuracy on multi-domain tasks and on multiple trustworthiness metrics, as depicted in \figref{fig:main}. Although all evaluations can be cast as classification after post-processing, quantization exhibits a different impact on trustworthiness metrics than on general-task accuracy. We analyze aggregated performance score and refusal rate across all metrics, thereby addressing the three questions in~\secref{sec:eval}.

\subsubsection{MMLU}

We evaluate the models on MMLU, which spans 57 subjects in a multiple-choice format to represent the performance in multi-domain tasks. All questions were answered, yielding a zero refusal rate. In terms of accuracy, the change relative to the 13B dense baseline is negligible for both 8-bit and 4-bit quantization. Further compressing the model to 3-bit leads to a noticeable performance drop; however, accuracy for most configurations remains higher than that of the 7B dense baseline. These results indicate that low-precision quantization can largely preserve multi-domain task accuracy compared to dense models and can outperform smaller-parameter dense models. Comparing the two quantization methods, both AWQ and GPTQ preserve accuracy at 8-bit, while AWQ is more robust at 4-bit and 3-bit.

\subsubsection{Adversarial Robustness}

Adversarial robustness is assessed on the AdvGLUE++ variants of three GLUE tasks (SST-2, QQP, and MNLI). At 8-bit, both AWQ and GPTQ achieve accuracy close to the 13B dense baseline. Under lower precisions, AWQ exhibits improved robustness at 4-bit and 3-bit, whereas GPTQ degrades by more than $10\%$, at 4-bit and 3-bit. Accuracy is computed over the valid label set; refusals and outputs that cannot be mapped to task labels are counted as errors. Consistent with this definition, we observe that AWQ low-bit models have low refusal rates, while GPTQ at higher compression yields more label-inconsistent (``contradictory'') answers rather than explicit abstentions, resulting in refusals. This pattern suggests that aggressive GPTQ compression may impair calibration and label consistency.

Comparing dense baselines, the 7B model outperforms the 13B model on AdvGLUE++ by $9.3\%$ and yields a lower refusal rate. One possible explanation is that smaller models are less sensitive to spurious lexical cues introduced by adversarial perturbations, whereas larger models are more easily misled by minor word changes. The observed correlation between lower refusal rates and higher accuracy holds for both dense and quantized variants.


\subsubsection{Fairness}

Fairness is assessed on the Adult dataset~\cite{adult_2}, with scores aggregated from two metrics: demographic parity difference (DPD) and equalized odds difference (EOD). Unlike the other trustworthiness metrics, the fairness score does not completely rely on correctly predicting the category. Both dense and quantized models achieve high fairness, with scores close to or above $80\%$ (higher is better under our normalized fairness score); all configurations also achieve low refusal rates, since the requests are not ambiguous or misleading. For AWQ, the fairness score remains similar at 8-bit and 4-bit, while at 3-bit the performance decreases; GPTQ attains high fairness at 8-bit and 3-bit, with a decrease at 4-bit. Additionally, the breakdown results show that the magnitude of DPD is typically lower than that of EOD, indicating that EOD contributes more to the overall unfairness; for high-fairness configurations, the difference between EOD and DPD is small.

Comparing dense baselines, the 7B model also outperforms the 13B model, as well as the quantized 13B variants. The performance gap is mainly associated with cases showing extreme base-rate imbalance, where the label distribution is highly skewed. In such cases, larger models appear more sensitive to the imbalance and make more biased classifications.

\subsubsection{Machine Ethics}

Machine ethics is assessed on the ETHICS dataset~\cite{hendrycksaligning}, reporting accuracy together with a false positive ratio (FPR) that penalizes false positive responses to jailbreaking and evasive inputs. At 8-bit, both quantization methods achieve performance similar to the 13B dense baseline. At lower bit-widths (4- and 3-bit), AWQ shows improved performance relative to the dense baseline, whereas GPTQ exhibits a significant drop. A breakdown indicates that the true positive rate on the immoral class is similar for all models; the 4- and 3-bit AWQ-quantized models benefit from a lower FPR on evasive sentences, while low-bit GPTQ-quantized models frequently produce invalid or abstaining outputs under few-shot, yielding a higher refusal rate. This highlights the model's degradation of in-context learning capability. 

Comparing dense baselines, the 7B dense model scores lower than the 13B dense model and lags behind the AWQ-quantized variants. Although the 7B model’s accuracy is approximately $10\%$ lower, its lower refusal rate narrows the gap with the 13B dense model.

\subsubsection{Out-of-Distribution Robustness}

OOD robustness is assessed on both style-shifted inputs and out-of-scope knowledge queries. For style-shifted inputs (e.g., Shakespearean phrasing), robustness means maintaining high accuracy with a low refusal rate. For out-of-scope queries, robustness requires detecting that the query is beyond the model’s knowledge and abstaining appropriately, while answering in-scope queries correctly, achieving high meaningful accuracy (MACC). AWQ-quantized models maintain OOD performance relative to the 13B dense baseline, whereas GPTQ-quantized models degrade significantly at 4- and 3-bit. A breakdown attributes the 3-bit degradation primarily to elevated refusal rates, and the 4-bit degradation to low accuracy on out-of-scope knowledge tasks, indicating a failure mode similar to that observed in the machine-ethics dimension.

Comparing dense baselines, the 7B model outperforms the 13B model and the quantized variants on OOD robustness. Specifically, in the zero-shot out-of-scope evaluation, the 7B model achieves nearly double the accuracy while exhibiting a higher refusal rate, indicating a more conservative behavior on knowledge-unknown queries.

\subsubsection{Observations and Insights}

The above observations provide insights into the three questions in~\secref{sec:eval}. \textit{1)} Quantization methods have heterogeneous impacts on multi-domain tasks and on different trustworthiness metrics. This difference is often missed by standard evaluations of quantized models. \textit{2)} For both AWQ and GPTQ, 8-bit quantization largely preserves performance. AWQ is more robust at 4- and 3-bit, whereas low-precision GPTQ can substantially degrade few-shot adherence and in-context learning in the LLaMA-2 series. \textit{3)} Smaller models, whether quantized or trained with fewer parameters, can outperform on certain trustworthiness dimensions by being less sensitive to ambiguous phrasing, where larger models are more likely to err. These insights motivate our attempt to enhance quantized models' robustness in trustworthiness, as described in ~\secref{sec:ensemble}.
\begin{figure*}[!ht]
  \centering
  \begin{minipage}[t]{0.55\linewidth}
    \centering
    \includegraphics[width=\linewidth]{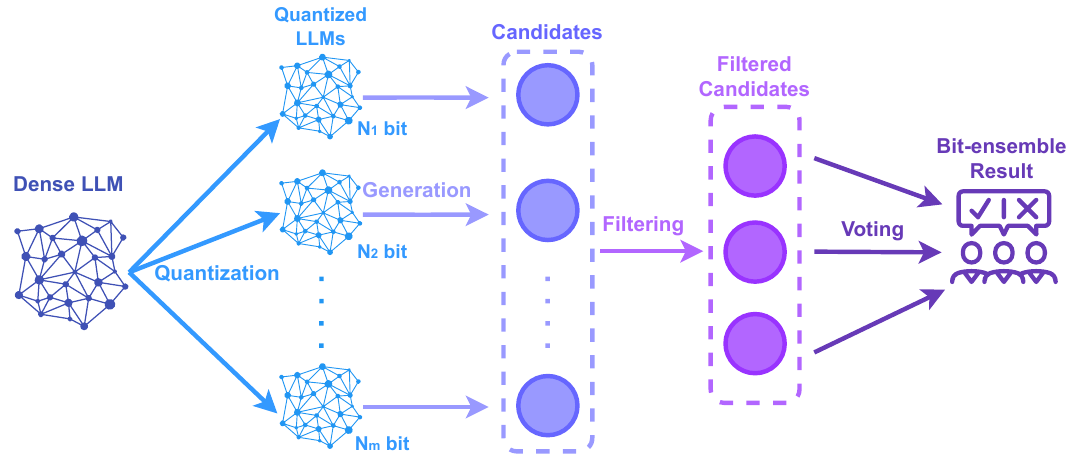}
    \captionof{figure}{Workflow of precision-ensemble voting. A dense LLM is quantized to multiple precisions; each quantized model generates its own response. After response filtering, the remaining responses are aggregated via unweighted majority voting.}\label{fig:ensemble}
  \end{minipage}\hspace{0.02\linewidth}%
  \begin{minipage}[t]{0.4\linewidth}
    \centering
    \includegraphics[width=\linewidth]{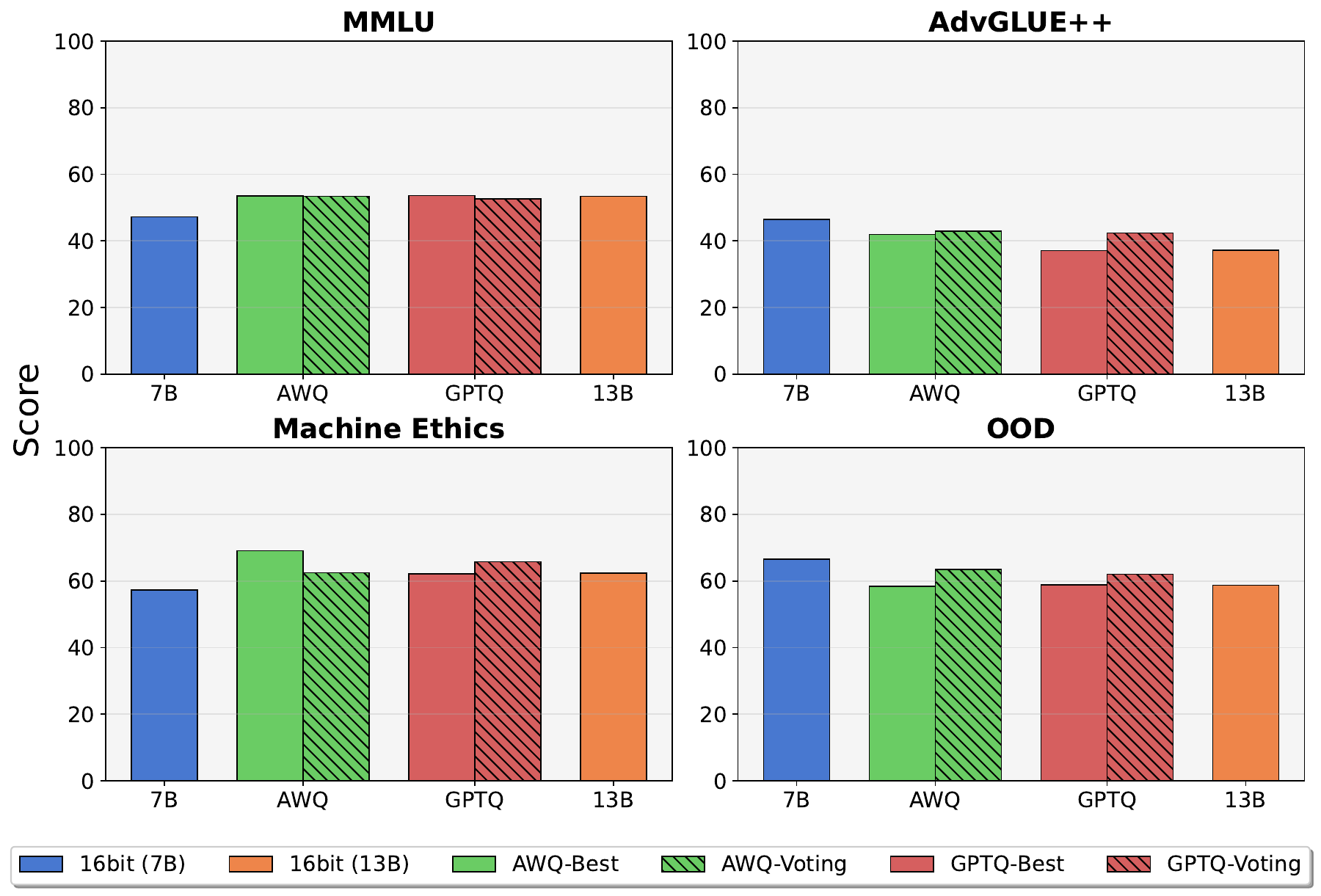}
    \captionof{figure}{The precision-ensemble voting mechanism maintains MMLU accuracy while consistently improving performance on the trustworthiness benchmarks.}\label{fig:voting}
  \end{minipage}
  \vspace{-15pt}
\end{figure*}

\section{Enhancing Trustworthiness Using Precision-Ensemble Voting} \label{sec:ensemble}

\subsection{Motivation}

In \secref{sec:eval}, we examine the performance of quantized models at different compression ratios on multi-domain tasks and trustworthiness metrics. As shown in \figref{fig:main}, models quantized to low bit-widths can surpass dense models on certain trustworthiness metrics while maintaining comparable accuracy on multi-domain tasks. However, a critical bottleneck of low-precision quantization is reduced robustness: quantized models are more vulnerable to high refusal rates and can suffer abrupt performance drops in specific scenarios. Prior work~\cite{hong2024comptrust} also reports performance instability across model families and quantization methods. These observations motivate mechanisms that enable low-precision models to provide robust and consistent predictions.

We address this bottleneck by leveraging the idea of test-time optimization~\cite{snell2024scaling, chen2025rethinking}, which invests additional inference compute to enhance performance. To this end, we propose a simple voting-based precision-ensemble that aggregates the predictions of multi-precision variants quantized from the same dense LLM.

\vspace{-5pt}

\subsection{Precision-Ensemble Voting}

We pursue robust quantized models via \textbf{precision-ensemble voting}, illustrated in \figref{fig:ensemble}. The procedure comprises four stages, quantization, generation, filtering, and voting, as detailed in Algorithm~\ref{alg:voting}. In the quantization stage (lines~1--3), the dense backbone (e.g., 13B) is quantized to multiple bit-widths or the same bit-width using different seeds. In the generation stage (lines 4--5), each quantized LLM generates a response in parallel, and predictions are mapped to discrete labels. Many trustworthiness benchmarks can be cast as classification, enabling the use of voting to aggregate results. Labels can be extracted using heuristics or an LLM-as-a-Judge~\cite{li2025generation}.

In the filtering stage (lines~6--9), we discard invalid or contradictory outputs (e.g., empty, unparseable, or multi-label generations) and mark refusals. For standard classification metrics, refusals are removed. For OOD robustness tasks that evaluate out-of-scope queries, explicit refusals (e.g., ``I don't know'') are retained as a dedicated label. In the final voting stage, the remaining candidates are aggregated via unweighted majority voting. In the event of a tie, we select the highest-precision model's prediction.

\begin{algorithm}[ht]
\caption{Precision-ensemble voting}
\begin{algorithmic}[1]
\Require Dense model $\mathcal{M}_{\mathrm{fp}}$, precisions $\mathcal{B}=\{N_1,\dots,N_m\}$,
         prompt $x$, decoding params $\theta$, refusal filter $f$
\Ensure Final label $\hat{y}$
\Statex
\ForAll{$N_i \in \mathcal{B}$}
  \State $\mathcal{M}_i \gets \mathrm{Quantize}(\mathcal{M}_{\mathrm{fp}}, N_i)$
\EndFor
\State \textbf{Generate:} \quad $G \gets \{\mathrm{Generate}(\mathcal{M}_i, x;\theta)\}_{i=1}^m$

\State \textbf{Map to labels:} \quad $C \gets \{\mathrm{Postprocess}(g)\mid g\in G\}$

\State \textbf{Filter refused candidates:} \quad $C' \gets \{c\in C \mid \neg f(c)\}$
\If{$C'=\emptyset$}
  \State \Return \textsc{Refused}
\EndIf

\State \textbf{Aggregation:} $\hat{y} \gets \mathrm{Maj.Voting}(C')$

\State \Return $\hat{y}$

\end{algorithmic}
\label{alg:voting}
\end{algorithm}

\vspace{-8pt}

\subsection{Effectiveness of Precision-Ensemble Voting}

\subsubsection{Evaluation Setup}

We quantize the dense LLaMA-2-13B-Chat model to three precisions: 3-, 4-, and 8-bit. Each quantized model and the precision-ensemble are evaluated on multi-domain tasks (MMLU) and on three trustworthiness metrics: adversarial robustness, machine ethics, and out-of-distribution robustness. The fairness dimension is excluded in this evaluation because it quantifies inter-group disparities between predictions rather than absolute accuracy, and is therefore not directly comparable under this setup. For each metric, we compare the ensemble prediction against the 7B and 13B dense baselines and against the best-performing single-precision result for each quantization method.

\subsubsection{Effectiveness on Multi-Domain Tasks}

According to \figref{fig:voting}, precision-ensemble voting achieves accuracy that is close to both the best single-precision results and the 13B dense baseline, and it outperforms the 7B dense model. The gains on MMLU are modest because low-bit quantization already preserves accuracy for these models. Nevertheless, the ensemble effectively mitigates instability observed in low-precision GPTQ-quantized models. 

\subsubsection{Effectiveness on Trustworthiness Metrics}

\figref{fig:voting} shows that precision-ensemble voting delivers robust, desirable performance. Compared with dense baselines, the ensemble results consistently outperform the 13B dense model by up to $5.8\%$, but fail to match the 7B dense baseline, as small models often benefit from being less sensitive to spurious lexical cues. Relative to the best single-precision results, the precision-ensemble voting approach performs better in most trustworthiness metrics and for both quantization frameworks.

We attribute these gains to two factors: \textit{1)} the filtering stage (for standard classification metrics) removes refusals, thereby lowering the measured refusal rate and preserving meaningful labels for aggregation; and \textit{2)} unweighted majority voting improves stability by reducing variance across bit-widths: when a quantized model fails on a specific instance, other bit-widths do not tend to fail as well, and the aggregation exploits this partial error diversity.

\vspace{-5pt}
\section{Challenges and Opportunities}

\subsection{Mixed Precision for Multi-Modal Trustworthiness}
\textit{Challenges:} The rapid development and increasing deployment of multi-modal LLMs, integrating vision, speech, action, and language, introduces a more complex trustworthiness problem than in text-only models. In particular, embodied AI systems, which rely on cross-modal reasoning to interact with the real world, demand heightened attention to trustworthiness across all modalities.

\textit{Opportunities:} Multi-modal setting opens new opportunities for modality-aware mixed-precision quantization that jointly considers efficiency and trustworthiness.
Different bit-ensemble strategies and precision scheduling can be adaptively tuned based on modality-specific information collected at runtime. Our findings in the single-modality case lay the groundwork for future exploration in multi-modal scenarios.

\vspace{-3pt}

\subsection{Joint Compression and Trust-Aware Optimization}
\textit{Challenges:} While mixed-precision quantization has proven effective in preserving or even enhancing trustworthiness, real-world deployments often combine it with other compression techniques such as pruning, tensor decomposition, and knowledge distillation. However, the interplay between these methods and their joint effect on trustworthiness remains underexplored. In particular, naive combinations may introduce unpredictable degradation in ethical behavior, adversarial robustness, or fairness, especially under low-bit regimes.

\textit{Opportunities:} 
We identify a significant opportunity to unify compression design with trustworthiness objectives. One promising direction is to formulate model compression as an automated sparsity search problem, where trustworthiness metrics are directly embedded in the optimization objective. This would enable the development of trust-aware auto-compression pipelines capable of jointly tuning bit-width, sparsity, and decomposition rank for maximal efficiency while enhancing trustworthiness.

\subsection{Algorithm-System-Hardware Co-Design}
\textit{Challenges:} Efficient system and hardware support for mixed-precision and bit-ensemble execution is critical for real-world deployment. At the hardware level, designing a unified compute unit that can efficiently and concurrently support operations at multiple precisions remains a major challenge, due to inherent trade-offs among performance, power, and area (PPA).
At the system level, effective scheduling and execution policies are required to manage precision diversity, particularly in multi-batch and multi-tenant deployment settings.

\textit{Opportunities:} 
The hardware and system design choices span a large configuration space. When jointly considered with algorithmic parameters, this forms a vast co-design space where algorithmic performance and hardware/system efficiency can be optimized together~\cite{9743481}.
This joint optimization can be tackled using techniques such as Bayesian optimization or reinforcement learning to efficiently explore the co-design space and derive Pareto-optimal solutions.
\section{Conclusion}

This work analyzes the impact of quantization on trustworthiness, highlights the stability bottleneck with low precision, and proposes a precision-ensemble voting approach to improve robustness. We underscore the importance of considering trustworthiness when compressing LLMs for deployment, using weight quantization as a representative case study with quantitative experimental results.
Future directions include extending trustworthiness analysis to multi-modal settings, exploring joint compression and trust-aware optimization frameworks, and pursuing end-to-end algorithm–system–hardware co-design for trustworthy and efficient LLM deployment.

\bibliographystyle{IEEEtran}
\bibliography{refs}

\end{document}